\pdfoutput=1
\documentclass[11pt]{article}

\usepackage{acl}
\usepackage{enumitem, array}

\usepackage{graphicx}
\usepackage{times}
\usepackage{latexsym}
\usepackage{natbib}
\usepackage{amsmath}
\usepackage{hyperref}

\usepackage{lipsum}
\usepackage{booktabs} % For better-looking tables

\usepackage{multirow}
\usepackage[T1]{fontenc}
\usepackage[utf8]{inputenc}

\usepackage{microtype}

\usepackage{inconsolata}

\title{Measuring Adversarial Datasets}

\author{
    \textbf{Yuanchen Bai}\thanks{~~equal contribution.} \quad
    \textbf{Raoyi Huang}$^{*}$ \\
    \textbf{Vijay Viswanathan}  \quad
    \textbf{Tzu-Sheng Kuo} \quad
    \textbf{Tongshuang Wu}\\
    Carnegie Mellon University, Pittsburgh PA, USA\\
    \small 
    \texttt{ybai2@andrew.cmu.edu} \quad \texttt{raoyih@andrew.cmu.edu} \\
    \small
    \texttt{vijayv@cs.cmu.edu} \quad 
    \texttt{tzushenk@cs.cmu.edu} \quad 
    \texttt{sherryw@cs.cmu.edu}
}

\begin{document}
\maketitle

\begin{abstract}
In the era of widespread public use of AI systems across various domains, ensuring adversarial robustness has become increasingly vital to maintain safety and prevent undesirable errors. Researchers have curated various adversarial datasets (through perturbations) for capturing model deficiencies that cannot be revealed in standard benchmark datasets. However, little is known about how these adversarial examples differ from the original data points, and there is still no methodology to measure the intended and unintended consequences of those adversarial transformations. In this research, we conducted a systematic survey of existing quantifiable metrics that describe text instances in NLP tasks, among dimensions of difficulty, diversity, and disagreement. 
We selected several current adversarial effect datasets and compared the distributions between the original and their adversarial counterparts. The results provide valuable insights into what makes these datasets more challenging from a metrics perspective and whether they align with underlying assumptions. 
\end{abstract}

\section{Introduction}

NLP models that exceed on standard benchmarks can exhibit unexpected errors when tested on different distributions~\citep{shenOutOfDistributionGeneralizationSurvey2021,ganinUnsupervisedDomainAdaptation2015,luoTakingCloserLook2019,blitzerDomainAdaptationStructural2006,ganinDomainAdversarialTrainingNeural2017}. 
In particular, a subcategory of distribution shift, \emph{adversarial attacks}, tends to cause severe safety concerns: models can be over- or under- sensitive to small changes on the input, and a seemingly reliable model may be easily broken under adversarial attack (e.g., large language models can be used to generate misinformation if we simply change a knowledge context in the prompt~\cite{qian2023merge}.

To flag such model behaviors, the research community has created an extensive list of adversarial datasets (and methods for generating them), e.g., TextFooler \citep{jinBERTReallyRobust2020}, Adversarial GLUE \citep{shenOutOfDistributionGeneralizationSurvey2021} and Adversarial NLI~\citep{nieAdversarialNLINew2020}.
Most of them are created by perturbing existing benchmark dataset throughs certain pre-defined operations (e.g., switching synonyms, adding or removing negations, changing the time).
Indeed, these datasets can effectively capture model vulnerability --- model performance tends to decrease significantly on this data compared to performance on the original benchmark.
Even ChatGPT\footnote{\url{https://openai.com/blog/chatgpt}}, a highly competent model by many measures, still lacks robustness to adversarial attacks~\cite{wangRobustnessChatGPTAdversarial2023,liuRobustnessTimeUnderstanding2023}.

%Models are usually evaluated on in-domain development datasets that follow a different distribution compared to e.g., deployment distribution. This distribution shift can lead to a degradation in model performance, presenting a challenge that numerous studies aim to address \citep{shenOutOfDistributionGeneralizationSurvey2021,ganinUnsupervisedDomainAdaptation2015,luoTakingCloserLook2019,blitzerDomainAdaptationStructural2006,ganinDomainAdversarialTrainingNeural2017}). 
%To ensure the reliability of the models, an evaluation dataset of higher quality is required. Among all the topics, adversalarity has gained increased attention in recent years. Pipelines and interfaces (e.g. \citealp{wallaceTrickMeIf2019}, GARD Project \footnote{\url{https://www.darpa.mil/program/guaranteeing-ai-robustness-against-deception}})are also designed to generate larger datasets and promote more convenient testing processes. 
%Various kinds of generative and perturbation-based attacks have been performed to construct more challenging datasets for model testing, such as TextFooler \citep{jinBERTReallyRobust2020}, Adversarial GLUE \citep{shenOutOfDistributionGeneralizationSurvey2021} and Adversarial NLI~\citep{nieAdversarialNLINew2020}. 
%As large language models (LLMs) become more prevalent and utilized in high-stake scenarios, their advesalarilty robustness is more thoroughly tested to ensure safe behaviors (\citep{wangRobustnessChatGPTAdversarial2023,liuRobustnessTimeUnderstanding2023}). However, as competent as ChatGPT is in various tasks, its advesalarilty robustness is still to be improved.

\begin{table*}[t]
\centering
\small

\begin{tabular}{r|p{0.2\textwidth}|p{0.2\textwidth}|l}
\toprule
\textbf{Category} & Diversity & Difficulty & Disagreement \\
\midrule
\textbf{Text-Level} &
Frequency Distribution\newline Redundancy\newline Keyword Extraction\newline Data Density\newline Topic Diversity/Density\newline Fairness-related Diversity 
& 
Lexical Complexity \newline Long Text Structure \newline Grammaticality \newline Text Coherence \newline Topic Distribution \newline  Usable Information \newline Confidence \newline Spurious Bias
& Label Disagreement \\
\midrule
\textbf{Annotation-Level} & Demographic Diversity & Subjective Ambiguity & Inter-Annotator \\
\bottomrule
\end{tabular}
\caption{Metrics Categories}
\label{tab:metricsCat}
\end{table*}

Despite its importance, there is currently no methodology to measure the intended and unintended consequences of adversarial transformations on a dataset, such as how those adversarial perturbations change the data distribution. This is partly caused by a lack of \emph{holistic methods for evaluating dataset properties} --- when we create and evaluate a dataset, what dimensions should we compare it with existing datasets?
While there exist ample surveys on metrics that capture model performances~\citep{liangHolisticEvaluationLanguage2022, leeEvaluatingHumanLanguageModel2022}, surveys on metrics for \emph{describing data} are more sparse and tend to focus on qualitative descriptions ~\cite[e.g.,][]{mitchellMeasuringData2022} than quantitative measures.
%Also, though various dimensions have been explored (e.g., ambiguity, implicit bias, not fluent, etc.), it’s not clear how they interact with each other, or how much they represent the distribution we care about. Thus, an integration of those metrics would be helpful for future research.
However, such quantification is important for contextualizing the usefulness of adversarial data.
If a method changes the distribution too aggressively (e.g., an over-emphasis on a particular domain or vocabulary that the model is not familiar with), evaluations on these datasets may be meaningless~\cite{bowman2021dangers}.

In this work, we explore quantitatively comparing adversarial data points (created through perturbation) with their original counterparts in a systematic way. 
We first conduct a systematic survey on existing quantifiable metrics that describe text data instances, broadly dividing them into three crucial dimensions: difficulty, diversity, and disagreement~\cite{mitchellMeasuringData2022}. We then run a subset of relevant metrics on two adversarial datasets, assessing the characteristics of these datasets and distribution patterns.

%generally focus on the following research questions:
%\begin{enumerate}[nosep, leftmargin=2.1em,labelwidth=*,align=left, label=Q.\arabic*]
%    \item What are the instance-level characteristics of the original and adversarial text pairs of each dataset?
%    \item Are there any particular distribution patterns? 
%    \item Do they align with the underlying assumptions mentioned in the papers describing the dataset?
%\end{enumerate}

% \sherry{We find that...}

We find that a comprehensive frame of meta-evaluation metrics, specifically focusing on difficulty, diversity, and disagreement, provides valuable insights for adversarial research. Through experimentation on our selected datasets, these metrics offer a detailed profiling of dataset attributes. A deeper exploration of these metrics at the characteristic level proves essential for tasks like dataset curation, model performance analysis, attack scenario design, and controlled experimental setups.

% Our work offers a comprehensive framework for evaluating text instances, enhancing our understanding of existing adversarial datasets, which will aid future evaluations of adversarial robustness and contribute to the development of more robust NLP models.

%To explore deeper into the topic, we want to explore whether there is any particular distribution that features those adversarial examples compared to those before changes are made. From this perspective, it is essential to find quantifiable metrics to describe each instance. However, in spite of comprehensive surveys about metrics for model performance, user experience, and data in general ( \citealp{liangHolisticEvaluationLanguage2022}, \citealp{leeEvaluatingHumanLanguageModel2022}, \citealp{mitchellMeasuringData2022} ), there still lacks a systematic review of the current existing metrics, especially quantifiable ones, for a particular data instance (basically text, e.g. paragraph, comment, sentence) in NLP tasks.

%This study provides an integration of quantifiable metrics deepening the understanding of text datasets on both instance-level and dataset-level, extending their usage in future research. Based on the metrics, current evaluation datasets of adversarial effect are further explored from their instance-level characteristics, helpful for future adversarial dataset construction and controlled experiment design.

\section{Metrics}
Inspired by \citet{mitchellMeasuringData2022}, we survey existing metrics on datasets from three dimensions: \emph{difficulty}, \emph{diversity}, and \emph{disagreement}. 
The challenges in this literature survey involve: distinguishing between dataset and text instance metrics, necessitating a meticulous examination of full-text content to identify metrics instead of relying solely on keyword searches, addressing the interdisciplinary nature of applicable metrics, and, as a result, not limiting them to specific paper venues. Thus, we apply the method of detecting anchor papers and snowball sampling on papers citing relevant literature. We finally compiled a set of 52 papers and several websites. The metrics are selected based on their widespread usage, state-of-the-art status, or high ranking in benchmarks. We then manually categorized the related metrics into the three dimensions above as shown in Table~\ref{tab:metricsCat}. The items listed in the table are middle-level categorizations and more details can be found Appendix~\ref{appendix:AppendixA-metrics}. Note that as shown in Table~\ref{tab:metricsCat}, these metrics are divided into \emph{text-} and \emph{annotator-} levels, as both instances and people who create them change the data distribution.

%Within each dimension, we further classify them into text-level and annotation-level metrics. Within each sub-group, we provide a discussion of several exemplary metrics and the methods that can be used to derive them. These examples are chosen based on their wide usage, state-of-the-art status, or high ranking in benchmarks. For detailed sub-grouping criteria and further illustrations, please refer to Appendix~\ref{sec:appendix}.

 %involves measuring data from various perspectives, including distance, density, diversity, tendency, and association. The paper provides multiple examples of data measurement metrics across different text and vision-related modalities.
 
% Inspired by the work, we concentrate specifically on text data, conducting a more comprehensive and in-depth exploration of existing literature. Furthermore, we investigate potential interconnections among diverse metrics and propose possible methods for combining them to gain valuable insights. Our approach follows a similar high-level categorization, which includes three essential dimensions: difficulty, diversity, and disagreement, as shown in Table~\ref{tab:metricsCat}.  Within each dimension, we further classify them into text-level and annotation-level metrics. Within each sub-group, we provide a discussion of several exemplary metrics and the methods that can be used to derive them. These examples are chosen based on their wide usage, state-of-the-art status, or high ranking in benchmarks. For detailed sub-grouping criteria and further illustrations, please refer to Appendix~\ref{sec:appendix}.

\begin{itemize}[nosep, leftmargin=1.2em,labelwidth=*,align=left]

\item \textbf{Diversity} characterizes \emph{the data variety in different aspects} such as topics and lexicons and granular levels such as samples or corpus. 
%It provides insights about data informativeness and coverage.
A more diverse dataset usually covers a wider range of topics and contexts that represent the complexity of real-world application scenarios better, thus evaluating the model generalizability \citep{gururanganAnnotationArtifactsNatural2018}. 
%Typical application scenarios include bias mitigation \citep{Bender2018DataSF} and robustness assurance \cite{Jia2017AdversarialEF} since various groups or perspectives are well represented, and a wider range of cases are tested for unexpected input handling with a more diverse dataset. 

\item \textbf{Difficulty} intuitively represents the hardness of learning an instance. It can be influenced by various factors, including whether a sentence is readable (e.g., lexical complexity~\citep{piresDefinitionLinguisticMetrics2017c} or text coherence~\cite{vadlapudiAutomatedEvaluationReadability2010}) and what it talks about (topic distribution); 
In the model testing context, it can also be reflected in certain model behaviors like confidence or whether any spurious bias is detected.

\item \textbf{Disagreement} refers to the variability of the assigned label \citep{swayamdiptaDatasetCartographyMapping2020}. It can emerge both among labelers, different models, and models' decisions among epochs. What is worth noticing is that labeler disagreements can provide crucial insights into data reliability and diversity issues \citep{aroyoDICESDatasetDiversity2023}.

\end{itemize}

The metrics we will discuss include the traditional ones and new ones and span over applied statistics, linguistics, and natural language processing. We are not aimed at an exhaustive list of metrics but to show the depth, breadth, and significance of them in texts and datasets inspection and their insights for enhancing model adversarial robustness.

\begin{table*}[t]
    \centering
    \small
    \begin{tabular}{l|l|l}
    \toprule
    \textbf{Dataset} & \texttt{LLM-Knowledge-Conflict}~\cite{xieAdaptiveChameleonStubborn2023} & \texttt{IMDB}~\cite{kaushik2019learning} \\
    \textbf{Attack} & Knowledge Conflict & Label Flipping\\
    \textbf{Task} & Entity-based QA & Sentiment Analysis\\
    \textbf{Comparison} & Parametric vs. Counter Memory & Original vs. Revised Reviews\\
    \bottomrule
    \end{tabular}
\caption{Experiment Datasets Overview}
\label{tab:dataset}
\end{table*}

\section{Experiment}
\paragraph{Dataset}
We picked two textual datasets that have pairs of original instances and perturbed ones for model evaluation:
%For dataset selection, we focus on test datasets related to NLP tasks and thus exclude adversarial datasets of images, etc. We extend our choice of datasets to all that exhibit adversarial effects, for instance, not only explicit adversarial datasets but also contrast datasets. 
%Furthermore, as examples whose perturbations are more focused on local changes will not reflect very clearly on our metrics, we direct our attention to datasets that involve high-level revisions.

%In light of the above considerations, our selection process for experimental datasets adheres to the following criteria: 1) The text datasets must consist of both the original test sets and revised sets, wherein the scope of revisions (attack scenarios) is clearly defined; 2) Models’ behaviors are evaluated on the attack scenarios and tend to have lower performance or a certain level of vulnerability; 3) The datasets should involve high-level revision instead of merely low-level perturbations.

\begin{itemize}[nosep, leftmargin=1.2em,labelwidth=*,align=left]
\item \texttt{LLM-knowledge-conflict}~\citep{xieAdaptiveChameleonStubborn2023}, a dataset that investigates the knowledge conflicts of LLMs. They construct alternative context paragraphs for question-answering tasks by perturbing an LLM-generated paragraph that represents the models' \emph{parametric memory}. We rely on its data points on StrategyQA generated using ChatGPT.
%“Counter-memory” (conflicting evidence) is constructed corresponding to the elicited “parametric-memory” (original internal knowledge) of the original datasets, POPQA and STRATEGYQA. Two versions of “Counter-memory” are generated by ChatGPT and GPT4 relatively for each dataset. 
%Considering the homogeneity in the dataset generation process, we chose one of them for further analysis. Thus, We run metrics on the memory pairs of POPQA(ChatGPT).

\item \texttt{IMDB Sentiment Contrast Set}~\cite{kaushik2019learning}, a dataset constructed by having crowdworkers manually edit IMDB movie reviews for flipping their labels with minimal perturbation. It is one of the contrast sets generated in the paper based on the concept of models' decision boundaries and has been proven to be more challenging for the models. 
%As explicit labels are provided, we can also calculate PVI and assess spurious bias in addition to the general text metrics.

\end{itemize}

Both datasets focus on higher-level perturbations; We believe they have a higher tendency to change the data distribution compared to lower-level local changes like synonym replacement.
%Table~\ref{tab:dataset} provides an overview of the datasets and selected metrics.

For both two datasets, we compute five metrics from the difficulty level (coherence, perplexity, FRE readability
score, semantic noise, and semantic clarity), four metrics from the diversity level (sentence, word, syllable and topic number) and one metric from the disagreement level (variance). For IMDB contrast set, we further compute PVI and assess spurious bias. The referred literature and computation methods are illustrated in the next subsection in detail.

\subsection{Metrics Implementation}
 We have carefully selected methods of calculating metrics across three dimensions (Table~\ref{tab:metricsImpA}, ~\ref{tab:metricsImpB}, ~\ref{tab:metricsImpC}), which can be found in Appendix~\ref{appendix:AppendixB-implementation} with more details. We have also implemented them as APIs to make them more convenient for other reference, released for the public in a GitHub repository\footnote{\url{https://github.com/BYC-Sophie/TextMetrics}}.

\section{Results}

\begin{figure*}
    \centering
    \includegraphics[width=0.8\linewidth]{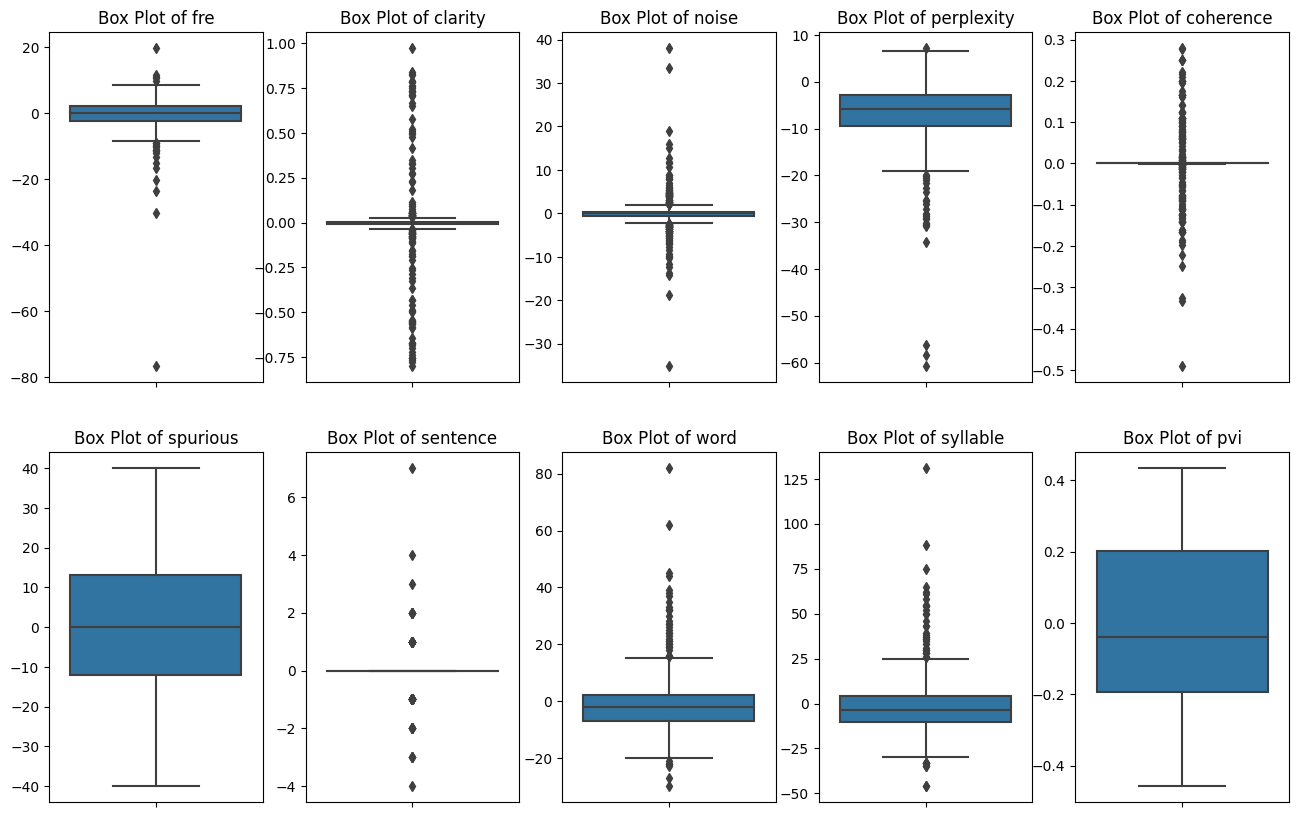}
    \centering
    \caption{Boxplot for instance-level differences ($d_{\text{ori}} - d_{\text{revised}}$) (IMDB)}
    \label{fig:boxplot-IMDB}
\end{figure*}

\begin{figure*}
    \centering
    \includegraphics[width=0.8\linewidth]{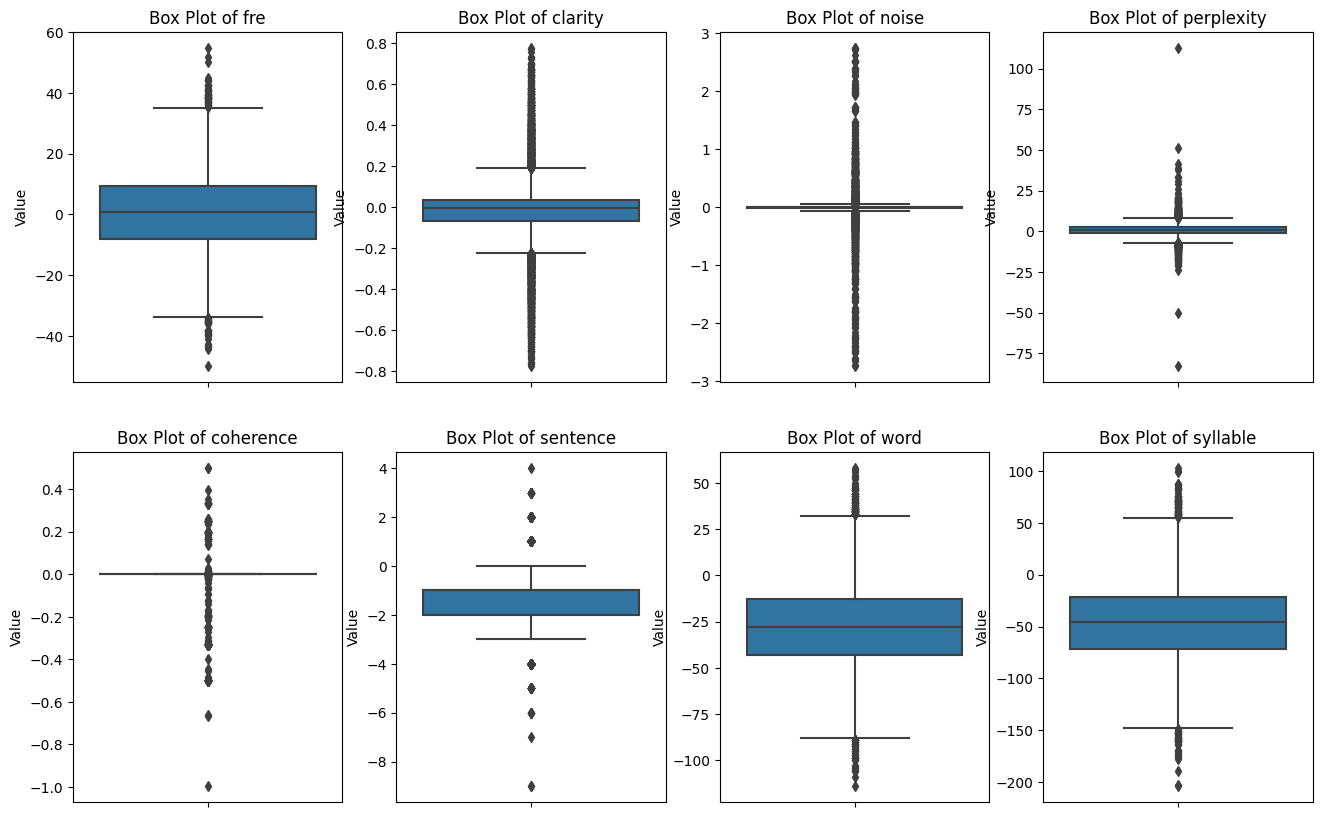}
    \centering
    \caption{Boxplot for instance-level differences ($d_{\text{ori}} - d_{\text{revised}}$) (LLM-Conflict)}
    \label{fig:boxplot-llm}
\end{figure*}

\paragraph{While IMDB datasets perturb mostly at the high level, Conflict-LLM datasets perturb both high-level and low-level textual information.} For the IMDB datasets, as the correlation in spurious bias and PVI turns very low, and more than half samples increase their perplexity, the revised datasets are mostly perturbed in fluency and task difficulty. In contrast, as the Conflict-LLM datasets show a low correlation in perplexity, coherence, sentence count, word count, and syllable count, and more than half of the samples increase their counts of sentences, words, and syllables, the revised datasets are perturbed in fluency and surface-level lexical structure (Table~\ref{tab:correlation}, Figure~\ref{fig:boxplot-IMDB}, Figure~\ref{fig:boxplot-llm}). We used Wilcoxon signed-rank test to test the significance of metrics differences and confirm our test results with one-sample t-test for metrics that roughly follow a normal distribution. While only perplexity, sentence count, word count, and syllable count exhibit a significant difference for IMDB dataset, all metrics differences are significant for LLM-Conflict datasets (significance level 0.025) (Table~\ref{tab:p-values-imdb}, Table~\ref{tab:p-values-llm}). The topic distributions after revision for two datasets show no apparent pattern, and the variance in labels makes no difference as there is only one label.

\paragraph{The values of coherence align with the papers' assumptions.} Regarding the papers' assumption on metrics, both datasets mention the metrics coherence. LLM-knowledge-conflict is stated to be generated by LLM to "ensure high coherence", and IMDb Sentiment Analysis is stated to "remain coherent and factually consistent". The experiment results show that the coherence metrics remain consistently high with minimal differences after revision, which shows the intended control during the adversarial transformations.

% We focus on two aspects regarding RQ3: 1) Metrics assumption alignment 2) Difficulty analysis.

% \subsubsection*{Metrics assumption alignment}

\paragraph{Unintended impacts are demonstrated from metrics' perspective.} The two datasets do not deliberately consider metrics other than the coherence above. All metrics changes do not show apparent patterns, especially LLM-Conflict, focusing on conflict memory, which is generated by language models and is thus reasonable to demonstrate limited controls for other metrics. However, there are still some changes induced by the transformation, but not always make the datasets more difficult: For IMDb, for example, a slight increase in perplexity indicates smaller fluency, while the increase in PVI suggests a slight decrease in difficulty. There is also a reduction in spurious bias in the dataset (Table ~\ref{tab:spurious_rounds}), which, in turn, can be interpreted as an increase in difficulty.

\section{Discussion}
\paragraph{Design diverse adversarial scenarios with transparency}
As mentioned in the dataset selection section, while there have been several adversarial datasets, such as HANS \citep{mccoyRightWrongReasons2019}, ANLI \citep{nieAdversarialNLINew2020} and SWAG \citep{zellersSWAGLargeScaleAdversarial2018}. However, most of them focus on the training process rather than evaluation and focus on low-level perturbations. Besides, although the adversarial version is available, it lacks a corresponding original version from which it was revised. This complicates comparisons and resembles the challenges we encountered during data selection experiments.
Thus, to better study adversalarity topics, it is worthwhile to explore additional high-level adversarial attack scenarios and make the generating process more transparent, such as providing the (original, revised) pairs, which would greatly enhance their utility.

\paragraph{Clarify what else makes the dataset more challenging}
The adversarial datasets are often designed to focus on specific attack scenarios, and their effectiveness becomes evident through the observed reduction in model performance and the mistakes made by the models. By conducting additional instance-level metrics, we can delve into a more micro perspective and gain a clearer understanding of whether the attacks have affected specific metrics or introduced additional complexities that further challenge the models.

\paragraph{Curate adversarial evaluation datasets with focuses}
For adversarial evaluation dataset curation, it is possible to concentrate on instances with particular metrics. This approach enhances interpretability and alignment with real-world application scenarios, which can assist researchers in gaining deeper insights into the inherent properties of datasets and the related models, and also make it easier to cater to various practical needs.

\section{Conclusion \& Limitations}
In this paper, we presented a systematic survey of the meta-evaluation metrics that quantify the difficulty, diversity, and disagreement of datasets to assist research in adversalarity. We experimented with two datasets LLM-knowledge-conflict and Contrast Set IMDb Sentiment Analysis, and used the calculated metrics to profile the dataset properties from different perspectives. Based on the analysis, we find it crucial to delve deeper into the metrics at the characteristic level, which can be helpful in various circumstances, such as dataset curation, analyzing model performance results, designing attack scenarios, and conducting controlled experiments. There are a few limitations to consider for future research. First, datasets that meet the criteria for local or high-level changes are rare, and there is a need for more diverse and transparent datasets in this regard. Second, the contrast dataset was published in 2020, during which BERT was used for testing and showed performance degradation. As more advanced structures and models appear, there is a need for more recent and challenging datasets to further explore these dynamics.

% It shows that the trends in spurious bias and PVI change much at the instance level, and differences in coherence and FRE are relatively significant for IMDb dataset. For LLM-Conflict dataset, the trends in FRE, perplexity, and coherence change much at the instance level, and the differences in FRE and noise are relatively significant. 

\bibliography{anthology,custom}

\appendix

\section{Metrics Related Work}
\label{appendix:AppendixA-metrics}

This is a section for the details of the criteria for the metrics classification.

\subsection{Diversity}
Diversity characterizes the data variety in different aspects such as topics and lexicons and granular levels such as samples or corpus. It provides insights about data informativeness and coverage. A more diverse dataset usually covers a wider range of topics and contexts that represent the complexity of real-world application scenarios better, thus evaluating the model generalizability \citep{gururanganAnnotationArtifactsNatural2018}. Typical application scenarios include bias mitigation \citep{Bender2018DataSF} and robustness ensurance \cite{Jia2017AdversarialEF} since various groups or perspectives are well represented, and a wider range of cases are tested for unexpected input handling with a more diverse dataset.

\subsubsection{Lexical diversity}
Lexical diversity measures the dispersion of tokens at different granular levels such as lexicons and n-grams from a statistical perspective rather than related to a specific NLP task. N-gram represents contiguous sequences of n items from texts where items can be phonemes, syllables, letters, or words depending on the application \citep{shannon1948}, where terms are most extensively used with Hidden Markov Models \citep{rabiner1989} or Language Models \citep{brown2020language} in areas such as text prediction and spelling correction. 

\begin{itemize}
    \item \textbf{Frequency distribution}: The most straightforward way to measure the lexical diversity is to calculate the number unique instances and their corresponding frequencies at different granular levels \citep{Zipf49}. Text data with higher lexical diversity should include more distinct tokens with a more evenly frequency distribution.

   \item \textbf{Redundancy}: Shannon entropy from information theory can measure the randomness and uncertainty of a text dataset at a granular level which is typically words or characters \citep{shannon1948}. A higher entropy represents less predictability, indicating less redundancy. Another metric is the generalized entropy index used primarily in economics to assess income distribution within a population \citep{theil1967economics}, which can be adapted to the text data context where individuals become tokens, and incomes become the corresponding frequencies. However, almost all those metrics do not capture the sequence order of tokens except n-grams, and n-grams are hard to use due to the exponential complexity of storing and retrieving them.

\end{itemize}

\subsubsection{Semantic diversity}
Semantic diversity captures the variation of key information in various aspects including keywords, topics, and contextual terms requiring special notice. While lexical diversity contains statistical measures at the surface level based on syntactical structure, semantic diversity intends to capture important information of the text data at a higher level of meaning.

\begin{itemize}
    \item \textbf{Keyword extraction}: Keyword extraction extracts vital terms from each text sample based on the term scores. A traditional metric to weigh term importance is TF-IDF \citep{SprckJones2021ASI}, which only uses the surface-level measures of term frequency and document frequency but was shown effective in the Information Retrieval area. However, TF-IDF only considers the useful terms after case conversion, stopword removal, and stemming without contextual and positional information. In addition to TF-IDF, RAKE (Rapid Automatic Keyword Extraction) \citep{Rose2010AutomaticKE}, TextRank \citep{Mihalcea2004TextRankBO}, YAKE (Yet Another Keyword Extractor) \citep{Campos2018YAKECA} are well-known models to extract keywords by surface-level knowledge, such as word frequencies, word co-occurrence frequencies, and sentence overlaps, as well as statistical manipulation. The keyword importance can also be extracted from term weights learned from large language models such as BERT \citep{devlinBERTPretrainingDeep2019}. 
    
    \item \textbf{Topic modeling}: Topic modeling to extract topics from the text data such as LDA (Latent Dirichlet Allocation) \citep{Blei2001LatentDA} that generate topic labels consisting of a group of keywords, which is one step further from plain keywords extraction methods. More advanced methods use architectures like graph-based models \citep{Kipf2016SemiSupervisedCW}, Neural Networks or transformers \citep{Vaswani2017AttentionIA}, learned text embeddings such as Word2Vec, Skip-gram \citep{Mikolov2013EfficientEO}, or GloVe \citep{Pennington2014GloVeGV} for topic modeling, and additional meta information. Improving on LDA, Gaussian LDA \citep{Das2015GaussianLF} uses Gaussian distributions instead of Multinomial distributions for topics with continuous word embeddings to better capture the word semantic relationships, and Hierarchical Dirichlet Process \citep{Teh2006AHB} uses a non-parametric Bayesian method to eliminate the need of the number of topics to be specified in advance. Also, some methods use Neural Networks such as TopicRNN \citep{Dieng2016TopicRNNAR} and graphs to model word-document relationships such as Graph-of-Words model \citep{Rousseau2015MainCR}. The state-of-the-art model BERTopic uses transformer-based language models such as BERT with a clustering algorithm to discover topics.

    \item \textbf{Data density}: As pre-trained text embeddings are naturally a vector in a multidimensional space, metrics like data density measure the vector density for each sample naturally. As data density increases, the average number of sample vector ending points within a unit space increases, which means the vectors become more similar, and the lexical diversity is likely to decrease. Data density at a more granular level terms can capture the linguistic density of the dataset. Also, the machine learning model KNN \citep{Cover1967NearestNP} can be used to calculate the average KNN density for each centroid to measure the vector density at a deeper level with consideration of clusters. The variation of data density can be an indicator of noise due to the occurrences of outliers.

    \item \textbf{Fairness-related diversity}: Some fairness-related diversity may need special attention such as gender and race. Several methods can be adopted to measure fairness-related diversity. The simplest method is to build a list of keywords for each minority group such as female and male, and parse the text samples to label the minority groups captured and corresponding frequencies \citep{Dixon2018MeasuringAM}. Samples with higher fairness-related diversity should include more diversity groups with higher frequencies. Also, samples whose contexts do not relate to fairness should be filtered out to avoid excessive calculation. The state-of-the-art methods are based on Transformers to tag each term with the corresponding fairness-related entity, which is adapted from the Name Entity Recognition task \citep{Sang2003IntroductionTT}.

    \item \textbf{Labeler demographic diversity}: Each labeler can be represented by a vector depending on the self-reported information by one-hot encoding or label encoding based on the feature properties, for example, a categorical variable is nominal or ordinal. For instance, females and males can be converted to a binary number 0 or 1, and labeler incomes can be labeled by the plain or normalized income values. The labeler demographic diversity can be calculated by average distance metrics from two labeler vectors such as Euclidean Distance \citep{DrEng2001PatternC}.
\end{itemize}

\subsection{Difficulty}

\subsubsection{Lexical complexity}
We use “lexical complexity” for surface-level features of a text, such as the number of complex words used or syntactic structure. These phonological and lexical units and patterns can provide intuitive indicators of the difficulty of a text \citep{piresDefinitionLinguisticMetrics2017c} and are widely used to create various metrics. Prominent examples include the Flesch Reading Ease (FRE), Fog Index (FOG) score, and Lesch-Kincaid Grade Level (FKGL) index, which are composite measures derived from combinations of these attributes \citep{AutomatedTextSimplification}. These indices are useful for determining whether a text is appropriate for a particular reading age, assuming that the texts are well-written, coherent, and relevant in content. \citep{siddharthanSyntacticSimplificationText2004}

\subsubsection{Text structure}
Explicit signal words and structure information (e.g. text-level, indent, order, connective) can impact the conjunction and cohesion of text, thus affecting the difficulty of understanding it (\citealp{siddharthanSyntacticSimplificationText2006}, \citealp{meyerComparativeSignalingGenerated2018a}). In the context of the widespread adoption of Large Language Models (LLMs), the importance of text structure also becomes more pronounced. For instance, in the online course "ChatGPT Prompt Engineering for Developers" offered by DeepLearning.AI in collaboration with OpenAI, the official recommendation highlights the significance of clarifying the structure to enhance the quality of generated responses (e.g. “Use delimiters to clearly indicate distinct parts of the input”)\footnote{\url{https://learn.deeplearning.ai/chatgpt-prompt-eng/}}.

\subsubsection{Grammaticality}
Grammatical errors can lead to confusion and misinterpretation of a text, increasing the difficulty of comprehending its intended meaning. Grammatical Error Correction (GEC)  has gained great attention in recent years. Various GEC models are trained and tested on benchmarks like CoNLL-2014, JFLEG,  BEA Shared Task, etc. (GEC work tracking\footnote{\url{http://nlpprogress.com/english/grammatical_error_correction.html}}
) GPT models also have a certain degree of good correction ability at the sentence level \citep{schmidt2023chatgpt}.  Multiple metrics can be utilized to quantify the efforts needed to correct the texts (e.g. Edit distance, Sentence similarity (\citealp{ferreiraAssessingSentenceSimilarity2016}, \citealp{parkComparisonEvaluationMetrics2020b})), which can be a proxy of grammaticality evaluation. 

\subsubsection{Text coherence}
The concept of coherence, which pertains to the semantic meaningfulness of text, is a crucial aspect of instance difficulty \citep{vadlapudiAutomatedEvaluationReadability2010}. Bert, a language model specifically pre-trained on the Next Sentence Prediction (NSP) task  \citep{devlinBERTPretrainingDeep2019}, has proven to be an efficient tool in quantifying text coherence metrics \citep{bommasaniIntrinsicEvaluationSummarization2020}. These metrics including the calculation of the average probability of each subsequent sentence conditioned on the preceding one, are proven to provide valuable insights into the overall coherence of multi-sentence texts \citep{bommasaniIntrinsicEvaluationSummarization2020}. Perplexity, one of the most widely used metrics for language model evaluation can also indicate how complex the instance seems to the model\footnote{\url{https://huggingface.co/docs/transformers/perplexity}} \citep{jelinekPerplexityMeasureDifficulty2005}. An instance with a lower perplexity value is considered more "probable" or "expected" by the language model, while higher perplexity indicates that the language model is less confident in predicting the sentence and may struggle to make accurate predictions for it.

\subsubsection{Topic distribution}
In addition to the number of identified topics, the topic distribution within a text plays a significant role in understanding its difficulty. This concept of topic distribution encompasses three important aspects, as discussed in \citealp{leePushingTextReadability2021a}: semantic richness, semantic clarity, and semantic noise. Semantic richness is measured through various indicators such as semantic neighbors, feature numbers, and contextual dispersion \citep{pexmanThereAreMany2008a}. In \citealp{leePushingTextReadability2021a}, a novel approach is introduced that quantifies semantic richness based on the product total of the sorted probabilities list (SPL) derived from topic modeling models. Additionally, a more intricate semantic structure within a text tends to increase its level of difficulty \citep{piresDefinitionLinguisticMetrics2017c}. Semantic clarity refers to the difficulty in identifying the main topic of the text and can be measured using the trend of the SPL \citep{leePushingTextReadability2021a}. On the other hand, semantic noise indicates the presence of less probable topics and can be assessed by examining the “tailness” of the SPL \citep{leePushingTextReadability2021a}.

\subsubsection{Usable information}
$\mathcal{V}$-Usable Information, extended from the work in information theory \citep{xuTheoryUsableInformation2020}, is proven to be effective in indicating the difficulty of a dataset \citep{ethayarajhUnderstandingDatasetDifficulty2022a}. It can be interpreted as “how much information an input variable X can provide about Y when constrained to functions V” \citep{chenCurriculumBroadCoverageBenchmark2022}. Pointwise V-information (PVI) can further indicate the difficulty of an individual instance for a given distribution  \citep{ethayarajhUnderstandingDatasetDifficulty2022a}. The higher the values, the easier the dataset/instance for the model/the distribution  \citep{ethayarajhUnderstandingDatasetDifficulty2022a}. These metrics are widely recognized in academic papers and serve as a reliable indicator (\citealp{chenCurriculumBroadCoverageBenchmark2022}, \citealp{trevisoEfficientMethodsNatural2022}). Other metrics regarding information include REV (Rationale Evaluation with conditional V-information), indicating “the amount of new, label-relevant information in a rationale beyond the information already available in the input or the label” \citep{chenREVInformationTheoreticEvaluation2022}.

\subsubsection{Confidence}
Confidence here refers to the degree of certainty the model possesses when making predictions about a given instance. It indicates the likelihood of a prediction being correct or a generated result satisfying the users. The precise meaning of confidence varies depending on the specific context. For example, it may represent the probability of making accurate predictions across epochs \citep{swayamdiptaDatasetCartographyMapping2020}, or the level of certainty that an answer will match a user's query in a Q\&A system\footnote{\url{https://learn.microsoft.com/en-us/}}. Intuitively, when the model is less confident, it suggests that the instance is more challenging to handle.

\subsubsection{Incidental bias (Annotation artifacts)}
Models may “solve the problems right, but for wrong reasons” due to annotation artifacts, resulting in an overestimation of their true machine commonsense capabilities (e.g. \citealp{gururanganAnnotationArtifactsNatural2018}, \citealp{gevaAreWeModeling2019}). This can lead to significant performance degradation when faced with adversarial or out-of-distribution samples. To mitigate this issue, for example, the AFLITE algorithm has been proposed as an effective means to detect dataset-specific spurious bias (\citealp{sakaguchiWinoGrandeAdversarialWinograd2020}, \citealp{brasAdversarialFiltersDataset2020}). The filtering process utilizes the Predictability Score, which measures the ratio of correct label predictions in each iteration, as a key metric. After going through more rounds of iterations, it the less likely for the remaining dataset to contain spurious bias, and can thus be more challenging. Pointwise Mutual Information (PMI) based filtering is another popular debias method \citep{gururanganAnnotationArtifactsNatural2018}. By summing all token-level PMI values for a text, instance pairs hard to discriminate can be detected (\citealp{gururanganAnnotationArtifactsNatural2018}, \citealp{sakaguchiWinoGrandeAdversarialWinograd2020}).

\subsubsection{Safety - Detected toxicity}
We also employ the term “Toxicty” as a broad category encompassing interconnected concepts such as hate speech and violent speech as in \citep{liangHolisticEvaluationLanguage2022}. Toxicity detection models can provide scores of various aspects. We do not assume the accuracy and reliability of those models. Thus, we refer to this metric as “detected toxicity” by the models utilized. Widely used toxicity detection tools include Perspective API\footnote{\url{https://perspectiveapi.com/}} \citep{leesNewGenerationPerspective2022}). This metric can help better evaluate a model’s performance on instances with various toxicity levels.

\subsubsection{Safety - Adversariality}
Prior research has emphasized the development of adversarial datasets as a means to enhance the robustness of models (e.g. \citealp{bartoloBeatAIInvestigating2020}, \citealp{kielaDynabenchRethinkingBenchmarking2021}, \citealp{dinanBuildItBreak2019}). With the widespread utilization of generative AI, there is an increasing concern regarding potentially unsafe behaviors that may give rise to undesirable issues, such as the inclusion of "violent, sexually explicit, or biased and derogatory stereotypes" \citep{parrishAdversarialNibblerDataCentric2023}. Current ongoing investigations are specifically targeting these challenging examples, particularly focusing on "adversarial attacks that can bypass existing safety filters" \citep{parrishAdversarialNibblerDataCentric2023}. By incorporating adversarial examples into evaluation datasets, the safety of models can be further examined and assessed.

\subsubsection{Annotator subjective ambiguity}
Annotations inherently carry the subjectiveness of the annotators. In certain situations, annotators may experience uncertainty when applying the pre-defined annotation criteria \citep{aroyoDICESDatasetDiversity2023}. We refer to this uncertainty as the "subjective ambiguity" of the annotators.

\subsection{Disagreement}

In human-labeled datasets, instances often exhibit disagreement among annotators, which serves as a significant characteristic. This disagreement can arise from three primary sources: individual differences, stimulus characteristics, and context, as discussed in \citep{basileWeNeedConsider2021}. Analyzing the instances and annotator disagreement can be valuable in assessing data reliability and selecting a meaningful subset of data. For instance, by considering background information of annotators such as age, gender, and education level, examining rater disagreement across different groups, including intersectional groups, can provide valuable insights into safety concerns \citep{aroyoDICESDatasetDiversity2023}. To measure and evaluate inter-annotator reliability catering to different data types and usage scenarios, various metrics are available and can be adapted. These include percent agreement \citep{mchughInterraterReliabilityKappa2012a}, Cohen's Kappa \citep{mchughInterraterReliabilityKappa2012a}, Intraclass Correlation Coefficient \citep{kooGuidelineSelectingReporting2016}, and the $A_m$ Agreement Measure, which is useful when assigning texts to more than one class \citep{bhowmickAgreementMeasureDetermining2008}.

\section{Metrics Implementation}
\label{appendix:AppendixB-implementation}
Here is a detailed illustration of our selected metrics' implementation.

We used FRE (Flesch Reading Ease) \citep{AutomatedTextSimplification} to evaluate the readability, which indicates the lexical complexity of an instance. To access spurious bias using the AFLITE algorithm, which has shown significant effectiveness. Hyperparameters are set proportionally to those stated in \citep{brasAdversarialFiltersDataset2020}, which have proven to be effective. We use the rounds of elimination as our metric, where more iterations lead to less potential for the remaining dataset to contain spurious bias. Regarding spurious bias, We conducted two sets of experiments using datasets, each containing 488 instances. We eliminated a total of 400 instances during each iteration, removing 10 instances at a time. This approach allowed us to track the elimination round for each instance. We then applied a threshold of 0.75 to identify instances with potentially higher spurious bias, determining the number of iterations needed to filter out those surpassing this specific threshold and the total removed instance number. To measure text coherence, we calculate semantic coherence \citep{bommasaniIntrinsicEvaluationSummarization2020}. This metric predicts the probability of each successive sentence conditioned on the previous one using BERT, indicating fluency at the sentence level. However, semantic coherence does not cover cases with only a single sentence. To address this, we include the perplexity score based on GPT2-large \footnote{\url{https://huggingface.co/gpt2-large}}, a widely recognized causal language model \citep{radfordLanguageModelsAre2019}. For assessing usable information, we use PVI, a commonly used indicator for instance/dataset distribution \citep{ethayarajhUnderstandingDatasetDifficulty2022a}. For topic distribution, we consider semantic clarity and noise. The calculation methods for these two are relatively alternative to the standard kurtosis equation from statistics and the kurtosis equation under Fisher's definition, as adopted and proved effective in \citealp{leePushingTextReadability2021a}.

We have also implemented them as APIs to make them more convenient for other reference and use. We have already set default values for the hyperparameters and default models corresponding to the calculation method. Nevertheless, users have the flexibility to fine-tune them to achieve more desirable results. It is worth noting that some metrics are label-independent, while others depend on label information, so users must provide sufficient details for the calculation process. Comprehensive explanations for each metric can be found in the GitHub repository\footnote{\url{https://github.com/BYC-Sophie/TextMetrics}}.

\begin{table*}[t]
\centering
\small
\begin{tabular}{@{}l |l |l @{}}
\toprule
\textbf{Category} & \textbf{Metrics Name} & \textbf{Computation Methods} \\ \midrule
Lexical Complexity & Readability & FRE (Flesch Reading Ease) Score \\
\midrule
Spurious Bias &  & Elimination rounds through AFLITE algorithm \\
\midrule
\multirow{2}{*}{Text Coherence} & Semantic coherence & Average successive sentence probability by Bert \\
 & Perplexity & Perplexity score based on GPT2-Large  \\
 \midrule
Usabel Information & PVI & Additional knowledge gained from labels \\
\midrule
\multirow{2}{*}{Topic Distribution} & Semantic clarity & Alternative of standard skewness equation \\
 & Semantic noise & kurtosis equation under Fisher definition \\ \bottomrule
\end{tabular}
\centering
\caption{Difficulty: Metrics Implementation}
\label{tab:metricsImpA}
\end{table*}

\begin{table*}[t]
\centering
\small
\begin{tabular}{@{}l |l |l @{}}
\toprule
\textbf{Category} & \textbf{Metrics Name} & \textbf{Computation Methods} \\ \midrule
Lexical diversity & Sentence, Word, Syllable Counts & Tokenizers and Dictionary  \\
\midrule
Topic Density & Topic Number & BertTopic \\
\bottomrule
\end{tabular}
\centering
\caption{Diversity: Metrics Implementation}
\label{tab:metricsImpB}
\end{table*}

\begin{table*}[t]
\centering
\small
\begin{tabular}{@{}l |l |l @{}}
\toprule
\textbf{Category} & \textbf{Metrics Name} & \textbf{Computation Methods} \\ \midrule
Label Disagreement & Instance Label Disagreement & Variance  \\
\bottomrule
\end{tabular}
\centering
\caption{Disagreement: Metrics Implementation}
\label{tab:metricsImpC}
\end{table*}

\begin{table*}[t]
    \centering
    \small
    \begin{tabular}{c|c|c|c|c|c}
        \toprule
            \textbf{Datasets}& FRE&  clarity&  noise&  perplexity&  coherence\\  \midrule
            IMDB& .9484 &  .8390&  .9607&  .9094&  .6932\\ 
            LLM-Conflict& .2189 &  .6078&  .6864&  \textbf{.0738}&  \textbf{-.0086} \\
        \bottomrule
    \end{tabular}

    \vspace{0.1cm} 
        
    \begin{tabular}{c|c|c|c|c|c}
        \toprule
            \textbf{Datasets}& spurious bias& pvi & sent\_count& word\_count& syllable\_count\\  \midrule
            IMDB& \textbf{.0796}& \textbf{.0861}& .9836& .9853& .9844\\ 
            LLM-Conflict& -& -& \textbf{.0675}& \textbf{.0317} & \textbf{.0182} \\
        \bottomrule
    \end{tabular}
    \caption{Correlation Values for All Metrics between Original and Revised Datasets}
    \label{tab:correlation}
\end{table*}

% \begin{table*}[t]
%     \centering
%     \scriptsize
%     \begin{tabular}{c|c|c|c|c|c|c|c|c|c|c}
%     \toprule
%          \textbf{Datasets}& FRE&  clarity&  noise&  perplexity&  coherence&  spurious bias& pvi & sent\_count& word\_count& syllable\_count\\  \midrule
%          IMDB& .9484 &  .8390&  .9607&  .9094&  .6932&  \textbf{.0796}& \textbf{.0861}& .9836& .9853& .9844\\  \midrule
%          LLM-Conflict& .2189 &  .6078&  .6864&  \textbf{.0738}&  \textbf{-.0086}&  -& -& \textbf{.0675}& \textbf{.0317} & \textbf{.0182} \\ 
%     \bottomrule
%     \end{tabular}
%     \caption{Correlation Values for All Metrics between Original and Revised Datasets}
%     \label{tab:correlation}
% \end{table*}

\begin{table*}[t]
    \centering
    \small
    \begin{tabular}{c|c|c|c|c|c}
    \toprule
        \textbf{Tests} & FRE&  clarity&  noise&  perplexity&  coherence\\  \midrule
         Wilcoxon signed-rank test& .4374&  .1613&  .5052&  \textbf{.0000}&  .0375\\ 
         one-sample t-test& .0537&  .7204&  .8607&  -&  -\\ 
    \bottomrule
    \end{tabular}
    
    % \bigskip % Adds an empty line
    \vspace{0.1cm} 
    
    \begin{tabular}{c|c|c|c|c|c}
    \toprule
        \textbf{Tests} & spurious bias& pvi& sent\_count& word\_count& syllable\_count\\  \midrule
         Wilcoxon signed-rank test& .9748& .7057& \textbf{.0048}& \textbf{.0000}& \textbf{.0000}\\ 
         one-sample t-test& 1.0000& -& \textbf{.0243}& -& -\\ 
    \bottomrule
    \end{tabular}
    \caption{p-values of Instance-level Differences for All Metrics (IMDB)}
    \label{tab:p-values-imdb}
\end{table*}

% \begin{table*}[t]
%     \centering
%     \scriptsize
%     \begin{tabular}{c|c|c|c|c|c|c|c|c|c|c}
%     \toprule
%         \textbf{Tests} & FRE&  clarity&  noise&  perplexity&  coherence&  spurious bias& pvi& sent\_count& word\_count& syllable\_count\\  \midrule
%          Wilcoxon signed-rank test& .4374&  .1613&  .5052&  \textbf{.0000}&  .0375&  .9748& .7057& \textbf{.0048}& \textbf{.0000}& \textbf{.0000}\\ 
%          one-sample t-test& .0537&  .7204&  .8607&  -&  -&  1.0000& -& \textbf{.0243}& -& -\\ 
%     \bottomrule
%     \end{tabular}
%     \caption{p-values of Instance-level Differences for All Metrics (IMDB)}
%     \label{tab:p-values-imdb}
% \end{table*}

\begin{table*}[t]
    \centering
    \scriptsize
    \begin{tabular}{c|c|c|c|c|c|c|c|c}
    \toprule
        & FRE&  clarity&  noise&  perplexity&  coherence& sent\_count& word\_count& syllable\_count \\  \midrule
        Wilcoxon signed-rank test& \textbf{.0000}&  \textbf{.0000}&  \textbf{.0000}&  \textbf{.0000}& \textbf{.0000}& \textbf{.0000}& \textbf{.0000}& \textbf{.0000}\\     
        one-sample t-test& \textbf{.0000}& \textbf{.0000}& .0289& \textbf{.0000}& -&  -&  \textbf{.0000} &  \textbf{.0000}\\ 
    \bottomrule
    \end{tabular}
    \caption{p-values of Instance-level Differences for All Metrics (LLM-Conflict)}
    \label{tab:p-values-llm}
\end{table*}

\begin{table*}[t]
    \centering
    \scriptsize
    \begin{tabular}{c|c|c|c|c|c|c|c|c}
        \toprule
        & & run-1 & run-2 & run-3 & run-4 & run-5 & mean \\
        \midrule
        \multirow{2}{*}{\textbf{iteration num}} & original & 3 & 1 & 1 & 1 & 3 & \textbf{1.8} \\
        & revised & 1 & 1 & 1 & 1 & 1 & \textbf{1} \\
        \multirow{2}{*}{\textbf{removal instance num}} & original & 21 & 2 & 3 & 4 & 23 & \textbf{10.6} \\
        & revised & 0 & 2 & 0 & 0 & 2 & \textbf{0.8} \\
        \bottomrule
    \end{tabular}
    \caption{Iteration and Instance Removal Count for Five Runs of Spurious Bias Calculation (IMDB)}
    \label{tab:spurious_rounds}
\end{table*}

% \begin{table}[t]
%     \centering
%     \scriptsize
%     \begin{tabular}{p{0.5cm}|p{0.5cm}|p{0.5cm}|p{0.8cm}|p{0.9cm}|p{0.9cm}|p{0.7cm}}
%     \toprule
%          FRE&  clarity&  noise&  perplexity&  coherence\\  \midrule
%          1.74&  60.78&  68.64&  7.38&  -0.86\\ 
%     \bottomrule
%     \end{tabular}
%     \caption{Correlation values for all metrics in \% (Conflict-LLM)}
%     \label{tab:correlation-llm}
% \end{table}

% \begin{table*}[t]
% \centering
% \small
% \begin{tabular}{|c|c|c|c|c|c|c|c|}
% \hline
%  &  & \textbf{run-1} & \textbf{run-2} & \textbf{run-3} & \textbf{run-4} & \textbf{run-5} & \textbf{mean} \\ \hline
% \multirow{2}{*}{\textbf{iteration num}} & \textbf{original} & 3 & 1 & 1 & 1 & 3 & \textbf{1.8} \\ \cline{2-8} 
%  & \textbf{revised} & 1 & 1 & 1 & 1 & 1 & \textbf{1} \\ \hline
% \multirow{2}{*}{\textbf{removal instance num}} & \textbf{original} & 21 & 2 & 3 & 4 & 23 & \textbf{10.6} \\ \cline{2-8} 
%  & \textbf{revised} & 0 & 2 & 0 & 0 & 2 & \textbf{0.8} \\ \hline
% \end{tabular}
% \label{tab:spurious_rounds}
% \caption{Iteration Count and Total Removed Instances in Five Runs of Spurious Bias Calculation (IMDB)}
% \end{table*}

\end{document}